\documentclass[10pt, a4paper]{article}
\usepackage{lrec}
\usepackage{multibib}
\newcites{languageresource}{Language Resources}
\usepackage{graphicx}
\usepackage{tabularx}
\usepackage{soul}
\usepackage{epstopdf}
\usepackage[latin1]{inputenc}

\usepackage{arabtex}
\usepackage{utf8}
\setcode{utf8}

\usepackage[font=small,skip=0pt]{caption}

\usepackage{hyperref}
\usepackage{xstring}

\title{Build Fast and Accurate Lemmatization for Arabic}

\name{Hamdy Mubarak}

\address{QCRI, Hamad Bin Khalifa University (HBKU), Doha, Qatar \\
         hmubarak@hbku.edu.qa}

\abstract{
In this paper we describe the complexity of building a lemmatizer for Arabic which has a rich and complex derivational morphology, and we discuss the need for a fast and accurate lammatization to enhance Arabic Information Retrieval (IR) results. We also introduce a new data set that can be used to test lemmatization accuracy, and an efficient lemmatization algorithm that outperforms state-of-the-art Arabic lemmatization in terms of accuracy and speed. We share the data set and the code for public. \\
\newline  }

\begin{document}

\maketitleabstract

\section{Introduction}
Lemmatization is the process of finding the base form (or lemma) of a word by considering its inflected forms. Lemma is also called dictionary form, or citation form, and it refers to all words having the same meaning.\newline

Lemmatization is an important preprocessing step for many applications of text mining and question-answering systems, and researches in Arabic Information Retrieval (IR) systems show the need for representing Arabic words at lemma level for many applications, including keyphrase extraction \cite{shishtawy2009arabic} and machine translation \cite{dichy2003roots}. In addition, lemmatization provides a productive way to generate generic keywords for search engines (SE) or labels for concept maps \cite{plisson2004rule}.\newline

Word stem is that core part of the word that never changes even with morphological inflections; the part that remains after prefix and suffix removal. Sometimes the stem of the word is different than its lemma, for example the words: believe, believed, believing, and unbelievable share the stem (believ-), and have the normalized word form (believe) standing for the infinitive of the verb (believe).\newline

While stemming tries to remove prefixes and suffixes from words that appear with inflections in free text, lemmatization tries to replace word suffixes with (typically) different suffix to get its lemma.\newline

This extended abstract is organized as follows: Section \ref{sec:background} shows some complexities in building Arabic lemmatization, and surveys prior work on Arabic stemming and lemmatization; Section \ref{sec:data} introduces the dataset that we created to test lemmatization accuracy; Section \ref{sec:system} describes the algorithm of the system that we built and report results and error analysis in section \ref{sec:evaluation}; and Section \ref{sec:discussion} discusses the results and concludes the abstract.

\section{Background}
\label{sec:background}
Arabic is the largest Semitic language spoken by more than 400 million people. It's one of the six official languages in the United Nations, and the fifth most widely spoken language after Chinese, Spanish, English, and Hindi. Arabic has a very rich morphology, both derivational and inflectional. Generally, Arabic words are derived from a root that uses three or more consonants to define a broad meaning or concept, and they follow some templatic morphological patterns. By adding vowels, prefixes and suffixes to the root, word inflections are generated. For instance, the word \<وسيفتحون> (wsyftHwn)\footnote{Words are written in Arabic, transliterated using Buckwalter transliteration, and translated.} ``and they will open'' has the triliteral root \<فتح> (ftH), which has the basic meaning of opening, has prefixes \<وس> (ws) ``and will'', suffixes \<ون> (wn) ``they'', stem \<يفتح> (yftH) ``open'', and lemma \<فتح> (ftH) ``the concept of opening''.\newline 

IR systems typically cluster words together into groups according to three main levels: root, stem, or lemma. The root level is considered by many researchers in the IR field which leads to high recall but low precision due to language complexity, for example words \<كتب، مكتبة، كتاب> (ktb, mktbp, ktAb) ``wrote, library, book'' have the same root \<كتب> (ktb) with the basic meaning of writing, so searching for any of these words by root, yields getting the other words which may not be desirable for many users.\newline

Other researchers show the importance of using stem level for improving retrieval precision and recall as they capture semantic similarity between inflected words. However, in Arabic, stem patterns may not capture similar words having the same semantic meaning. For example, stem patterns for broken plurals are different from their singular patterns, e.g. the plural \<أقلام> (AqlAm) ``pens'' will not match the stem of its singular form \<قلم> (qlm) ``pen''. The same applies to many imperfect verbs that have different stem patterns than their perfect verbs, e.g. the verbs \<استطاع، يستطيع> (AstTAE, ystTyE) ``he could, he can'' will not match because they have different stems.  Indexing using lemmatization can enhance the performance of Arabic IR systems.\newline

A lot of work has been done in word stemming and lemmatization in different languages, for example the famous Porter stemmer for English, but for Arabic, there are few work has been done especially in lemmatization, and there is no open-source code and new testing data that can be used by other researchers for word lemmatization. Xerox Arabic Morphological Analysis and Generation \cite{beesley1996arabic} is one of the early Arabic stemmers, and it uses morphological rules to obtain stems for nouns and verbs by looking into a table of thousands of roots.\newline

Khoja's stemmer \cite{khoja1999stemming} and Buckwalter morphological analyzer \cite{buckwalter2002arabic} are other root-based analyzers and stemmers which use tables of valid combinations between prefixes and suffixes, prefixes and stems, and stems and suffixes. Recently, MADAMIRA \cite{pasha2014madamira} system has been evaluated using a blind testset (25K words for Modern Standard Arabic (MSA) selected from Penn Arabic Tree bank (PATB)), and the reported accuracy was 96.2\% as the percentage of words where the chosen analysis (provided by SAMA morphological analyzer \cite{graff2009standard}) has the correct lemma.\newline

In this paper, we present an open-source Java code to extract Arabic word lemmas, and a new publicly available testset for lemmatization allowing researches to evaluate using the same dataset that we used, and reproduce same experiments.

\section{Data Description}
\label{sec:data}
To make the annotated data publicly available, we selected 70 news articles from Arabic WikiNews site \url{https://ar.wikinews.org/wiki}. These articles cover recent news from year 2013 to year 2015 in multiple genres (politics, economics, health, science and technology, sports, arts, and culture.) Articles contain 18,300 words, and they are evenly distributed among these 7 genres with 10 articles per each.\newline

\begin{figure}[!h]
\begin{center}
\includegraphics[scale=0.45]{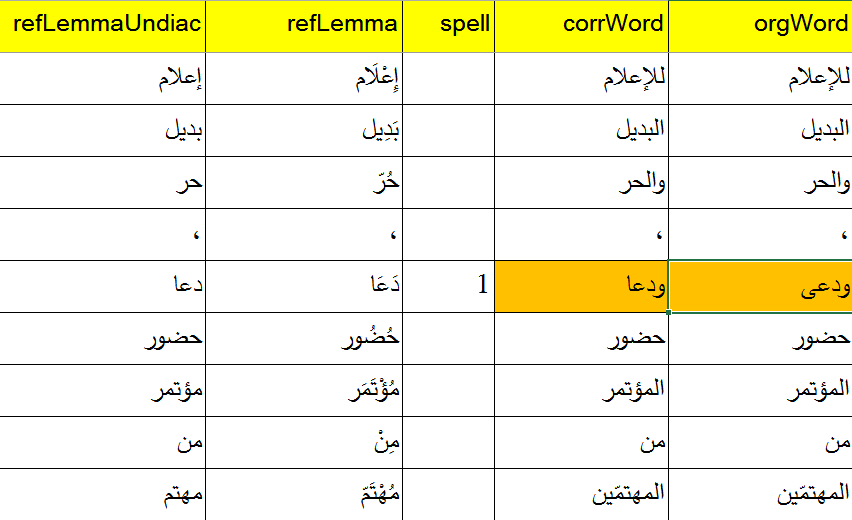} 
\caption{Lemmatization of WikiNews corpus}
\label{fig.WikiLemma}
\end{center}
\end{figure}

Word are white-space and punctuation separated, and some spelling errors are corrected (1.33\% of the total words) to have very clean test cases. Lemmatization is done by an expert Arabic linguist where spelling corrections are marked, and lemmas are provided with full diacritization as shown in Figure \ref{fig.WikiLemma}.\newline

As MSA is usually written  without diacritics and IR systems normally remove all diacritics from search queries and indexed data as a basic preprocessing step, so another column for undiacritized lemma is added and it's used for evaluating our lemmatizer and comparing with state-of-the-art system for lemmatization; MADAMIRA.

\section{system Description}
\label{sec:system}
We were inspired by the work done by \cite{darwish2016farasa} for segmenting Arabic words out of context. They achieved an accuracy of almost 99\%; slightly better than state-of-the-art system for segmentation (MADAMIRA) which considers surrounding context and many linguistic features. This system shows enhancements in both Machine Translation, and Information Retrieval tasks \cite{abdelali2016farasa}. This work can be considered as an extension to word segmentation.\newline

\begin{figure}[!h]
\begin{center}
\includegraphics[scale=0.38]{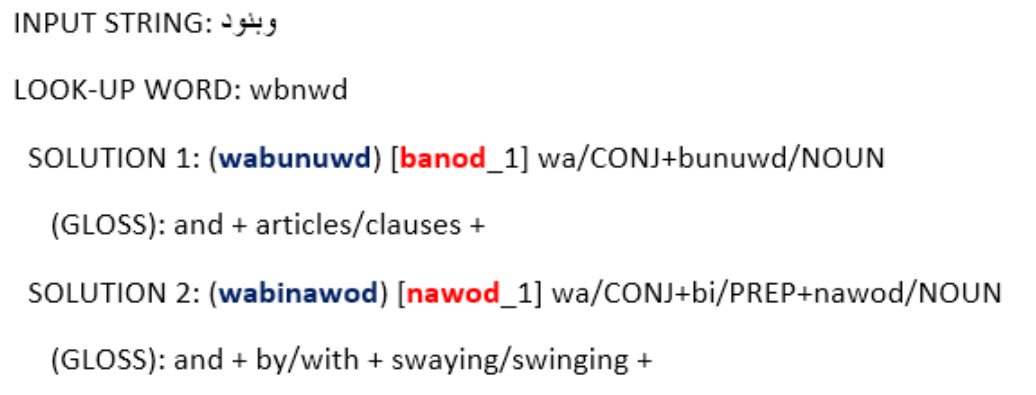} 
\caption{Buckwalter analysis (diacritization forms and lemmas are highlighted)}
\label{fig.buckwalter}
\end{center}
\end{figure}

From a large diacritized corpus, we constructed a dictionary of words and their possible diacritizations ordered by number of occurrences of each diacritized form. This diacritized corpus was created by a commercial vendor and contains 9.7 million words with almost 200K unique surface words. About 73\% of the corpus is in MSA and  covers  variety of genres like politics, economy, sports, society, etc. and the remaining part is mostly religious texts written in classical Arabic (CA). The effectiveness of using this corpus in building state-of-the-art diacritizer was proven in \cite{darwish2017diacritization}.For example, the word \<وبنود> (wbnwd) ``and items'' is found 4 times in this corpus with two full diacritization forms \<وَبُنُودِ، وَبُنُودٍ> (wabunudi, wabunudK) ``items, with different grammatical case endings'' which appeared 3 times and once respectively. All unique undiacritized words in this corpus were analyzed using Buckwalter morphological analyzer which gives all possible word diacritizations, and their segmentation, POS tag and lemma as shown in Figure \ref{fig.buckwalter}. 

The idea is to take the most frequent diacritized form for words appear in this corpus, and find the morphological analysis with highest matching score between its diacritized form and the corpus word. This means that we search for the most common diacritization of the word regardless of its surrounding context. In the above example, the first solution is preferred and hence its lemma \<بند> (banod, bnd after diacritics removal) ``item''.\newline

While comparing two diacritized forms from the corpus and Buckwalter analysis, special cases were applied to solve inconsistencies between the two diacritization schemas, for example while words are fully diacritized in the corpus, Buckwalter analysis gives diacritics without case ending (i.e. without context), and removes short vowels in some cases, for example before long vowels, and after the definite article \<ال> (Al) ``the'', etc.\newline

It worths mentioning that there are many cases in Buckwalter analysis where for the input word, there are two or more identical diacritizations with different lemmas, and the analyses of such words are provided without any meaningful order. For example the word \<سيارة> (syArp) ``car'' has two morphological analyses with different lemmas, namely \<سيار> (syAr) ``walker'', and \<سيارة> (syArp) ``car'' in this order while the second lemma is the most common one. To solve tis problem, all these words are reported and the top frequent words are revised and order of lemmas were changed according to actual usage in modern language.\newline

\begin{figure*}[!h]
\begin{center}
\includegraphics[scale=0.45]{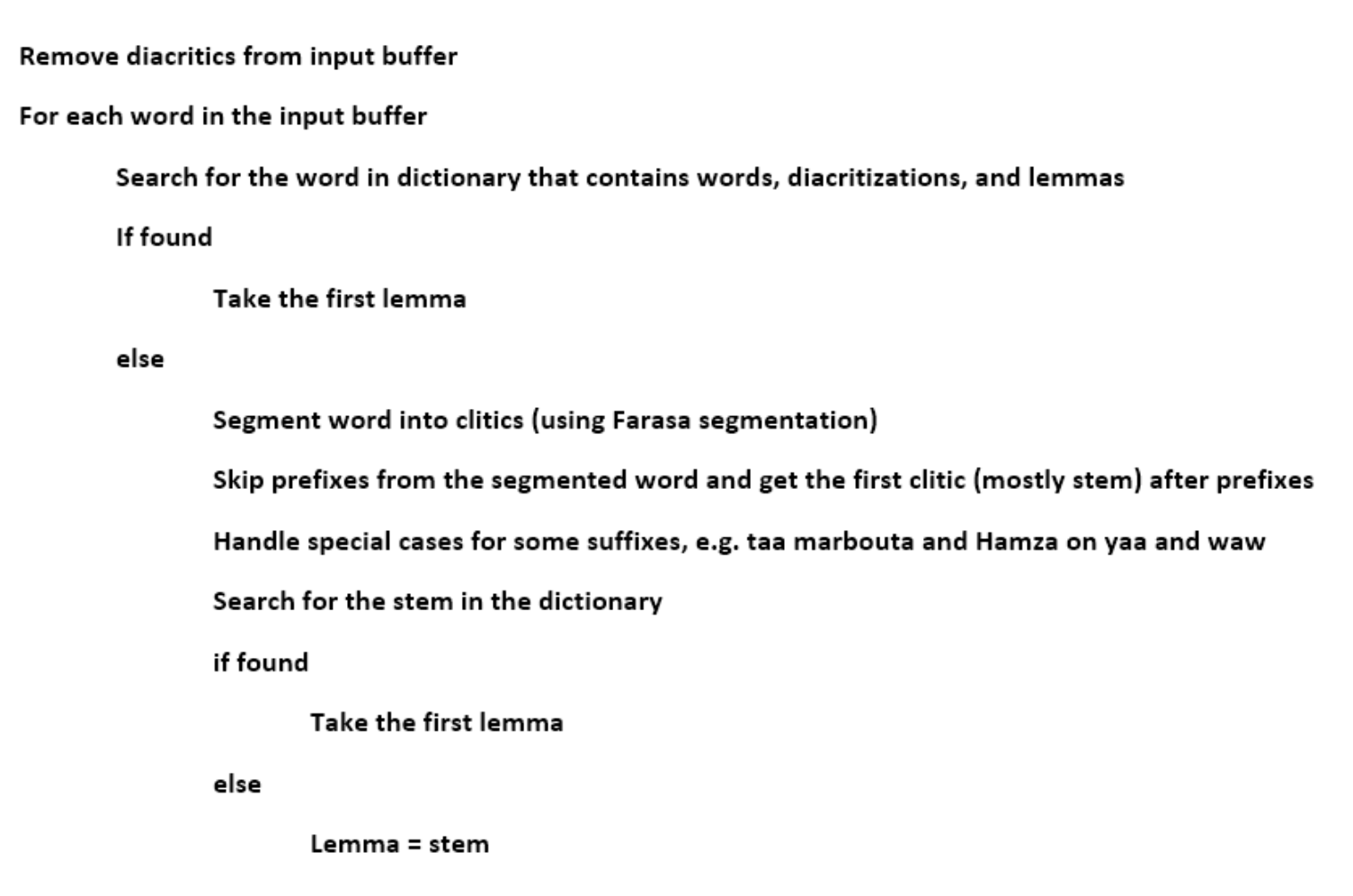} 
\caption{summary of lemmatization algorithm}
\label{fig.lemmatization}
\end{center}
\end{figure*}

The lemmatization algorithm can be summarized in Figure 
\ref{fig.lemmatization}, and the online system can be tested through the site \url{http://alt.qcri.org/farasa/segmenter.html}

\section{Evaluation}
\label{sec:evaluation}
Data was formatted in a plain text format where sentences are written in separate lines and words are separated by spaces, and the outputs of MADAMIRA and our system are compared against the undiacritized lemma for each word. For accurate results, all differences were revised  manually to accept cases that should not be counted as errors (different writings of foreign names entities for example as in \<هونغ كونغ، هونج كونج> (hwng kwng, hwnj kwnj)  ``Hong Kong'', or more than one accepted lemma for some function words, e.g the lemmas \<في، فيما> (fy, fymA) are both valid for the function word \<فيما> (fymA) ``while'').\newline

Table \ref{table:results} shows results of testing our system and MADAMIRA on the WikiNews testset (for undiacritized lemmas). Our approach gives +7\% relative gain above MADAMIRA in lemmatization task.

\begin{table}[ht]
\begin{center}
\begin{tabular}{|c|c|}
\hline
System & Accuracy \\ \hline
Our System & \textbf{97.32\%} \\
MADAMIRA & 96.61\% \\
\hline
\end{tabular}
\caption{Lemmatization accuracy using WikiNews testset}
\label{table:results}
\end{center}
\end{table}

In terms of speed, our system was able to lemmatize 7.4 million words on a personal laptop in almost 2 minutes compared to 2.5 hours for MADAMIRA, i.e. 75 times faster. The code is written entirely in Java without any external dependency which makes its integration in other systems quite simple.

\subsection{Error Analysis}
Most of the lemmatization errors in our system are due to fact that we use the most common diacritization of words without considering their contexts which cannot solve the ambiguity in cases like nouns and adjectives that share the same diacritization forms, for example the word \<أكاديمية> (AkAdymyp) can be either a noun and its lemma is \<أكاديمية> (AkAdymyp) ``academy'', or an adjective and its lemma is \<أكاديمي> (AkAdymy) ``academic''. Also for MADAMIRA, errors in selecting the correct Part-of-Speech (POS) for ambiguous words, and foreign named entities. \newline

In the full paper, we will quantify error cases in our lemmatizer and MADAMIRA and give examples for each case which can help in enhancing both systems.

\section{Discussion}
\label{sec:discussion}
In this paper, we introduce a new dataset for Arabic lemmatization and a very fast and accurate lemmatization algorithm that performs better than state-of-the art system; MADAMIRA. Both the dataset and the code will be publicly available. We show that to build an effective IR system for complex derivational languages like Arabic, there is a a big need for very fast and accurate lemmatization algorithms, and we show that this can be achieved by considering only the most frequent diacritized form for words and matching this form with the morphological analysis with highest similarity score. We plan to study the performance if the algorithm was modified to provide diacritized lemmas which can be useful for other applications.

\section{Bibliographical References}
\label{main:ref}

\bibliographystyle{lrec}
\bibliography{lrec2018}

\end{document}